\documentclass[lettersize,peerreviewca]{IEEEtran}
\usepackage{amsmath,amsfonts}
\usepackage{amssymb}
\usepackage{algorithm}
\usepackage{algorithmic}
\usepackage{array}
\usepackage[caption=false,font=normalsize,labelfont=sf,textfont=sf]{subfig}
\usepackage{textcomp}
\usepackage{stfloats}
\usepackage{url}
\usepackage{verbatim}
\usepackage{multirow}
\usepackage{graphicx}
\usepackage{cite}
\usepackage{siunitx}
\begin{document}

\title{Learning-Based Approximate Nonlinear Model Predictive Control Motion Cueing}

\author{Camilo Gonzalez Arango, Houshyar Asadi, Mohammad Reza Chalak Qazani, Chee Peng Lim
\thanks{Authors 1 and 2 are at Institute for Intelligent Systems Research and Innovation, Deakin University, Waurn Ponds, Victoria, 3216, Australia. Author 3 is at Sohar University, Sohar, 311, Oman. Author 4 is at Swinburne University, Hawthorn, Victoria, 3122, Australia.  }
\thanks{Manuscript submitted 28 Feb 2025}}

\markboth{}%
{Gonzalez \MakeLowercase{\textit{et al.}}: Learning-Based Approximate Nonlinear Model Predictive Control Motion Cueing}

\IEEEpubid{}

\maketitle

\begin{abstract}
Motion Cueing Algorithms (MCAs) encode the movement of simulated vehicles into movement that can be reproduced with a motion simulator to provide a realistic driving experience within the capabilities of the machine. This paper introduces a novel learning-based MCA for serial robot-based motion simulators. Building on the differentiable predictive control framework, the proposed method merges the advantages of Nonlinear Model Predictive Control (NMPC) — notably nonlinear constraint handling and accurate kinematic modeling — with the computational efficiency of machine learning. By shifting the computational burden to offline training, the new algorithm enables real-time operation at high control rates, thus overcoming the key challenge associated with NMPC-based motion cueing. The proposed MCA incorporates a nonlinear joint-space plant model and a policy network trained to mimic NMPC behavior while accounting for joint acceleration, velocity, and position limits. Simulation experiments across multiple motion cueing scenarios showed that the proposed algorithm performed on par with a state-of-the-art NMPC-based alternative in terms of motion cueing quality as quantified by the RMSE and correlation coefficient with respect to reference signals. However, the proposed algorithm was on average 400 times faster than the NMPC baseline. In addition, the algorithm successfully generalized to unseen operating conditions, including motion cueing scenarios on a different vehicle and real-time physics-based simulations. 
\end{abstract}

\IEEEpeerreviewmaketitle


\section{Introduction}
\begin{figure}[!b]
\centering
\includegraphics[width=5.5cm]{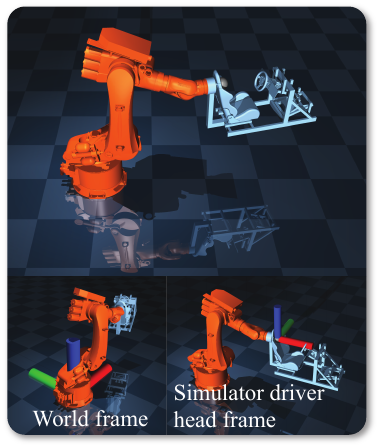}
\caption{MuJoCo model of motion simulator based on the KUKA KR500 robot arm. Bottom images show the reference frames relevant to motion cueing.}
\label{fig:motionSimulator}
\end{figure}
Motion Simulators (MSs) are machines that can emulate the experience of driving a simulated vehicle in a controlled and risk-free environment. These machines are frequently used in the aviation and automotive industries for applications such as driver training, product development and research. Figure \ref{fig:motionSimulator} shows the virtual development environment of a general purpose MS, where the motion platform is a serial robot. The scheme that controls how the motion platform moves is known as a Motion Cueing Algorithm (MCA). In short, this type of algorithm encodes the movement experienced aboard the simulated vehicle into movement that feels as realistic as possible and that can be reproduced with the motion platform given its kinematic and dynamic constraints.

The history of the motion cueing field can be summarised as transitions between four paradigms. The first paradigm was characterised by filter-based approaches and started with the Classical Washout Filter (CWF) \cite{conrad1970motion}. The CWF consists of high- and low-pass filters applied to the angular velocity and specific force signals measured aboard the simulated vehicle. These filtered signals are rendered via two mechanisms. High-frequency cues are integrated into displacement commands and rendered by rotations and translations of the motion platform. Alternatively, low-frequency cues are rendered via tilt coordination, a strategy that uses somatogravic illusion to simulate sustained linear acceleration via pitch or roll \cite{macneilage2007bayesian}. The use of high-pass filters in the CWF naturally introduces a tendency, known as washout, for motion platforms to return to their neutral position. However, because the time-invariant cutoff frequencies of the high- and low-pass filters control both motion fidelity and washout, tuning the CWF to achieve good performance under a wide range of operating conditions is difficult \cite{grant1997motion}. In practice, the algorithm is tuned to deliver acceptable performance for the expected worst-case scenario -- an approach that results in conservative cues for mid-range frequencies and which does not guarantee workspace boundary compliance. These shortcomings have been addressed in adaptive filtering methods, where cut-off frequencies are adjusted based on MS states and tracking error \cite{parrish1975coordinated}.

The second motion cueing paradigm was characterised by sensation-centered approaches that embed models of the Human Vestibular System (HVS) in the motion cueing pipeline. This type of approach was inspired by the idea that an MS should aim to reproduce sensed motion instead of the actual motion of the simulated vehicle. In short, the HVS is a set of organs located in the inner ear that sense angular velocity and linear acceleration \cite{asadi2017semicircular, asadi2016review}. Research on this organ system has yielded models from which the sensation response of an average human can be estimated. In addition, such studies have revealed the existence of sensation thresholds, below which, an average human cannot tell the difference between two different motion profiles. During the second paradigm, these results were integrated into filtering approaches to come up with algorithms that reproduce sensed motion. For example, Ariel and Sivan \cite{ariel1984false} extended the adaptive gradient descent washout filter by considering the HVS and aiming to minimise sensation error during cutoff frequency adjustments. Similarly, Telban and Cardullo \cite{telban2002nonlinear} further developed this type of approach by including visual-vestibular models in a nonlinear filtering scheme.

Beyond considering the HVS, the complexity and variety of filter adaptation schemes also increased during the second paradigm. For example, specific algorithms were proposed for the kinematics of hexapod and serial robot simulators with \cite{qazani2020adaptive} and without \cite{asadi2015incorporating} consideration of joint limits and other metrics such as dexterity \cite{qazani2021adaptive}. Nevertheless, all approaches remained filter-based and susceptible to disadvantages like overshooting during operation as a result of filter adaptation \cite{miunske2019new}, the introduction of lag due to filtering \cite{fang2010performance}, and an inherent limitation on the variety of motion cues that can be generated with fixed-order filters.

The third paradigm moved away from filtering approaches towards model-based control and established Model Predictive Control (MPC) as the state-of-the-art motion cueing method. Dagdelen et al.~\cite{dagdelen2009model} pioneered this approach with the first MPC-based MCA. This work demonstrated the key advantages of MPC such as the ability to seamlessly handle simulator limitations as optimisation constraints, the ability to generate cues accounting for future predictions of the simulator states, and the ease of tuning via a small set of weights. Since then, the adoption of MPC in motion cueing has grown steadily. Garrett and Best \cite{garrett2013model} proposed the use of a joint space plant model for hexapod motion platforms allowing to consider position and velocity constraints by using linear time varying MPC. Qazani et al.~\cite{qazani2019linear} extended this work by avoiding linearisations over the prediction horizon. Most notably, the adoption of Nonlinear MPC (NMPC) has enabled the use of joint space plants with constraints at the acceleration, velocity and position levels. This method has been successfully applied to serial robot \cite{katliar2018real}, cable driven \cite{katliar2017nonlinear}, and parallel motion platforms \cite{bruschetta2016nonlinear}. Overall, the consideration of nonlinearities in NMPC-based MCAs allows more efficient exploitation of the capabilities of a given motion platform by enabling the use of complex kinematic models and nonlinear constraints on all states, inputs and outputs. However, the computational cost of NMPC poses a significant challenge for real-time applications. In literature, this problem has been circumvented through small performance compromises such as using lower control rates \cite{katliar2017nonlinear}, considering short prediction horizons \cite{bruschetta2016nonlinear}, and downsampling the plant over the prediction horizon \cite{katliar2018real}.

Fueled by the progress in the artificial intelligence field, a fourth motion cueing paradigm has started to develop in recent years, characterised by the use of machine learning techniques as part of the motion cueing pipeline. Several works have focused on the use of neural networks (NNs) to predict the reference signals over the prediction horizon in MPC MCAs. Typically, reference signals are assumed constant over the horizon, however, it has been shown that better performance can be achieved when predictions from an NN trained to learn driving patterns are used \cite{mohammadi2016future, qazani2021prediction, qazani2021time}. Machine learning has also been applied to the computation cost problem of MPC. Rengifo et al.~\cite{rengifo2018solving} studied the use of a Hopfield NN as a means to replace the costly quadratic program solution involved in linear MPC. Their approach was demonstrated for a one degree-of-freedom MCA without tilt coordination. Similarly, Koyuncu et al.~\cite{koyuncu2020novel} trained a deep NN (DNN) to learn the outputs of an infinite horizon linear MPC MCA. By learning from examples of the cues generated for several drivers driving around a simulated track, they showed that an NN could replace the computationally expensive optimisation step in MPC. Most recently, Scheidel et al.~\cite{scheidel2024deep} proposed the first purely learning-based MCA using reinforcement learning. In their work, an agent learns what Cartesian movement to command next to achieve a suitable motion cue given a target motion profile and the present simulator states. The success of the approach was demonstrated on a lane change manoeuvre example where the learning-based algorithm was compared to an optimally tuned CWF. This type of learning-based approach is still in its infancy. The shortcomings of \cite{rengifo2018solving, koyuncu2020novel, scheidel2024deep} include the formulation of the problem in Cartesian space instead of joint space, the lack of consideration of nonlinearities such as those encountered in joint-space models of serial robot simulators, and the lack of comparisons against NMPC. 

Herein, we present a new learning-based MCA for serial robot simulators that belongs to the new paradigm and is rooted in the well-studied MPC approach. Our method uses Differentiable Predictive Control (DPC) \cite{drgovna2022differentiable}, a framework inspired by MPC, which solves the computational cost issue at inference time by shifting the cost to the offline training phase. Compared to existing learning-based MCAs, our approach uses a nonlinear joint space plant, which embeds the kinematics of the serial robot motion platform and accounts for joint acceleration, velocity and position limits. Moreover, our method benefits from some of the most useful features of MPC such as the ability to tune the controller using weights on the loss function, and the flexibility to enforce system limits as optimisation constraints. All while remaining real-time capable at high control rates. We demonstrate the effectiveness of this new MCA through simulation experiments and comparisons to exact NMPC. Results from these tests showcase the robustness and competitiveness of the new algorithm. To the best of our knowledge, this is the first approach of its kind in the motion cueing field. 

The remainder of the paper is organised as follows. Section \ref{sec:preliminaries} introduces DPC and the nonlinear motion cueing plant that we seek to control. Section \ref{sec:dpc_mca} introduces the plant and policy models that were evaluated and used during the development of the new MCA. Section \ref{sec:datasets} explains how we generated the datasets required for training and validation of the algorithm. Section \ref{sec:methods} presents the model evaluation and training methodologies. Section \ref{sec:results} contains all results and showcases the performance of the new MCA. Section \ref{sec:conclusion} summarises our findings and future research directions.

\section{Preliminaries}
\label{sec:preliminaries}

\subsection{Differentiable Predictive Control}
\begin{figure*}[!ht]
\centering
\includegraphics[width=\linewidth]{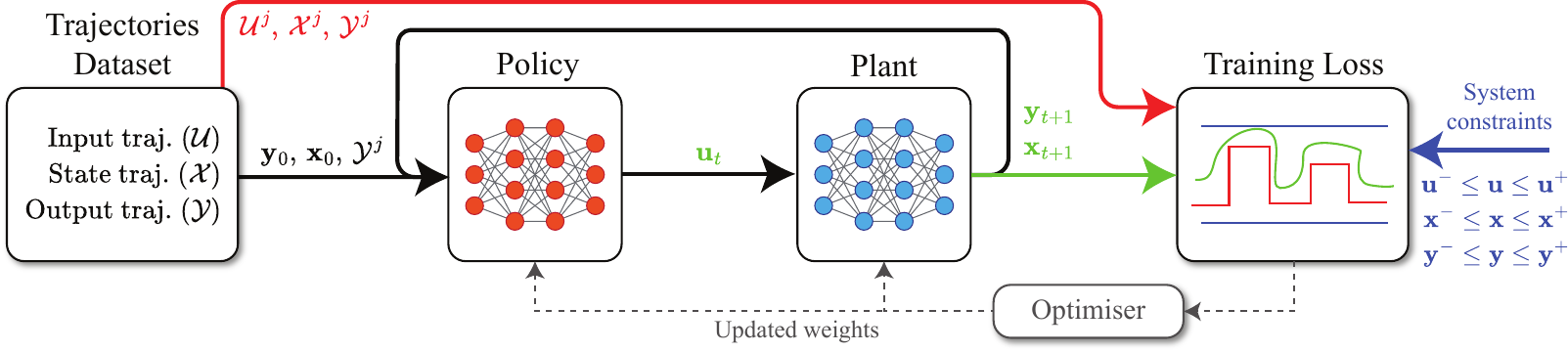}
\caption{Architecture and training setup for the DPC-based MCA.}
\label{fig:method_explanation}
\end{figure*}
Differentiable Predictive Control (DPC) is a learning-based control algorithm that is heavily inspired on MPC \cite{drgovna2022differentiable}. Typically, a DPC controller is made up of two models, the ``policy" and the ``plant" as shown in Figure \ref{fig:method_explanation}. The plant is a learnable function, such as a DNN, trained to learn the dynamics of the system being controlled. As in MPC, the plant in DPC is used to perform receding horizon predictions. Alternatively, the policy is a learnable function trained to decide which inputs to provide to the plant given the current state of the system and the desired future outputs. This part of the algorithm replaces the costly online optimisation step in MPC.

Training of a complete DPC solution happens in two stages. First, the plant model is trained in open loop mode, meaning that the policy block in Figure \ref{fig:method_explanation} is ignored and pre-recorded input signals ($\mathcal{U}$) are passed directly to the plant block. The predicted plant outputs ($\mathbf{y}$) and states ($\mathbf{x}$) are then compared against expected responses $(\mathcal{Y}, \, \mathcal{X})$. These prediction errors are used to form the training loss function. Next, the policy is trained using the closed-loop setup shown in Figure \ref{fig:method_explanation} including the pre-trained plant model. The loss function is designed with a similar format to the cost function in a typical MPC application, including terms for reference output tracking, control action minimisation, control smoothing, and terminal costs. At the beginning of training, the policy is randomly initialized and achieves poor regulation performance. In subsequent training steps, the weights of the policy are adjusted to minimise the MPC inspired loss and any other constraint-related losses. With enough training steps and data, the final result is a policy that behaves like MPC. Next, we introduce the analytical motion cueing plant that needs to be approximated for our proposed DPC-based MCA.

\subsection{Nonlinear Motion Cueing Plant Model}
\label{sec:analytical_plant}

The MCA proposed in this study is designed for the virtual MS model shown in Figure \ref{fig:motionSimulator} which is based on the KUKA KR500 robot arm. Table \ref{tab:robotKinematics} summarises the kinematic properties of the simulator. Using the well-proven approach by Katliar et al. \cite{katliar2018real}, we model the motion cueing process at the joint level, using the forward kinematics function of the robot and its first two derivatives. This formulation enables calculation of the angular velocity and tilt-coordinated linear acceleration experienced at the end effector given the joint positions $(\mathbf{q})$, velocities $(\dot{\mathbf{q}})$ and accelerations $(\ddot{\mathbf{q}})$. Put concisely, the motion cueing process can be modeled using equation set \eqref{eq:analytical_plant}.
\begin{subequations}
  \label{eq:analytical_plant}
  \begin{align}
    \label{eq:stateTransition}
     \begin{bmatrix} \Dot{\mathbf{q}} \\ \Ddot{\mathbf{q}}\end{bmatrix} & = \begin{bmatrix} 0 & \mathbf{I} \\ 0 & 0\end{bmatrix} \begin{bmatrix} \mathbf{q} \\ \Dot{\mathbf{q}}\end{bmatrix} + \begin{bmatrix} 0 \\ \mathbf{I}\end{bmatrix} \Ddot{\mathbf{q}}                                                   \\
    \label{eq:FK}
     \mathbb{F}(\mathbf{q})                                                                           & = \begin{bmatrix} \mathbf{R}_{H}^W & \mathbf{r} \\ \mathbf{0}       & 1\end{bmatrix} = \prod^6_{i=0} T_i^{i-1}(q_i) \, \cdot \, T^6_{H}                                                                                                               \\
    \label{eq:dFK}
     \mathbb{\Dot{F}}(\mathbf{q},\,\Dot{\mathbf{q}})                                                  & = \begin{bmatrix} \mathbf{\Dot{R}}_{H}^W & \mathbf{\Dot{r}} \\ \mathbf{0}             & 1\end{bmatrix}                             = \mathbb{J}(\mathbb{F}(\mathbf{q}))\,\Dot{\mathbf{q}}                       \\
    \label{eq:ddFK}
     \mathbb{\Ddot{F}}(\mathbf{q},\,\Dot{\mathbf{q}})                                                 & = \begin{bmatrix} \mathbf{\Ddot{R}}_{H}^W & \mathbf{\Ddot{r}}\\ \mathbf{0} & 1 \end{bmatrix} = \mathbb{J}(\mathbb{\Dot{F}}(\mathbf{q}))\,\Dot{\mathbf{q}} + \mathbb{J}(\mathbb{F}(\mathbf{q}))\,\Ddot{\mathbf{q}} \\
    \label{eq:rotationalMeasurements}
     \mathbf{\omega}_H                                                                                & = \mathbf{R}_{W}^H \, \mathbf{\Dot{R}}_{H}^W = \left(\mathbf{R}_{H}^W\right)^{\intercal} \,\mathbf{\Dot{R}}_{H}^W                                                                                                                                     \\
    \label{eq:linearMeasurements}
     \mathbf{a}_H                                                                                     & = \mathbf{R}_{W}^H \, \mathbf{\Ddot{r}}      = \left(\mathbf{R}_{H}^W\right)^{\intercal} \,\mathbf{\Ddot{r}}                                                                                                                                          \\
    \label{eq:tiltCoordination}
     \mathbf{a}_{H_{TC}}                                                                              & = \mathbf{a}_{H} + \left(\mathbf{R}_{W}^H \, \cdot \, -g^W\right)                                                                                                                                                                                     \\
    \label{eq:outputEquation}
     \mathbf{y}_t                                                                                     & = \left[\omega_{H_x} \;\; \omega_{H_y} \;\; \omega_{H_z} \;\; a_{H_{TC_x}} \;\; a_{H_{TC_y}} \;\; a_{H_{z}}\right]
  \end{align}
\end{subequations}

Equation \eqref{eq:stateTransition} is the state transition equation consisting of one double integrator per robot joint. The input of the system is $\ddot{\mathbf{q}}$ and the state vector is $[\mathbf{q}, \dot{\mathbf{q}}]$. Eq. \eqref{eq:FK} is the forward kinematics function of the robot $(\mathbb{F} \in \mathbb{R}^{4\times4})$, where each matrix $T_i^{i-1}(q_i)$ corresponds to the homogeneous transformation between subsequent joints, and $T^6_{H}$ corresponds to the transformation between the head frame and the last joint of the robot. Each of the matrices $T_i^{i-1}(q_i)$ is a function of the current joint position and the Denavit Hartenberg (DH) parameters of the joint and link. The matrix $\mathbf{R}_{H}^W \in \mathbb{R}^{3\times3}$ is the rotation from the head frame ($H$) to the world frame ($W$) and the vector $\mathbf{r} \in \mathbb{R}^3$ is the position of the head frame in the world frame. The first and second derivatives of $\mathbb{F}$ are give by \eqref{eq:dFK} and \eqref{eq:ddFK}, respectively -- these allow to calculate the angular velocity $\mathbf{\Dot{R}}_{H}^W$ and linear acceleration $\mathbf{\Ddot{r}}$ experienced at the end effector. Next, Eqs. \eqref{eq:rotationalMeasurements} and \eqref{eq:linearMeasurements} convert the inertial measurements from the world frame to the head frame and Eq. \eqref{eq:tiltCoordination}, applies tilt coordination to the linear acceleration measurements. The final outputs of the plant are grouped in \eqref{eq:outputEquation} which contains the angular velocity and tilt-coordinated linear acceleration measurements.
\begin{table}[!b]
\renewcommand{\arraystretch}{1.3}
\centering
\caption{Kinematic properties and Denavit Hartenberg parameters of the KR500 motion simulator}
\label{tab:robotKinematics}
\resizebox{\columnwidth}{!}{%
\begin{tabular}{lrrrrrr}
\textbf{Joint}     & \textbf{1} & \textbf{2} & \textbf{3} & \textbf{4} & \textbf{5} & \textbf{6} \\
\hline
$q^+ \, (rad)$*     & 3.229  & 1.920  & 0.942  & 6.109  & 2.094  & 6.109  \\
$q^- \, (rad)$*     & -3.229 & -0.698 & -3.316 & -6.109 & -2.094 & -6.109 \\
$\left|\Dot{q}\right|^+ \, (rad \cdot s^{-1})$*  & 1.571  & 1.396  & 1.309  & 1.571  & 1.449  & 2.269  \\
$\left|\Ddot{q}\right|^+ \, (rad  \cdot s^{-2})$* & 1.676  & 1.197  & 2.189  & 0.564  & 1.625  & 1.317  \\ \hline
DH a $(m)$       & 0.500  & 1.300  & -0.055 & 0.000  & 0.000  & 0.000  \\
DH d $(m)$      & 1.045  & 0.000  & 0.000  & 1.025  & 0.000  & 0.290  \\
DH alpha $(rad)$   & -1.571 & 0.000  & -1.571 & 1.571  & -1.571 & 0.000  \\
DH offset $(rad)$ & 0.000  & -1.571 & 0.000  & 0.000  & 0.000  & 0.000 \\ \hline
\multicolumn{7}{l}{*The superscripts $^+$ and $^-$ indicate upper and lower limits respectively.}
\end{tabular}%
}
\end{table}

\section{Learing-Based Motion Cueing Algorithm}
\label{sec:dpc_mca}
In this section we introduce the model architectures that were considered and tested during the development of the new MCA. The plant approximation problem was split into two parts, namely the state transition approximation and the output approximation problems. In the analytical model \eqref{eq:analytical_plant}, the state transition step is given by \eqref{eq:stateTransition}, while the output step involves equations \eqref{eq:FK} -- \eqref{eq:outputEquation}. The motivation for splitting the problem stemmed from the sequential nature of the equations. At each time step, the outputs of the state transition equation become the inputs of the output equations. Therefore, state prediction errors propagate through the output model and can make it difficult to identify if poor performance of a particular output model is due to poor state predictions or due to the output model itself. In addition, dividing the problem into small modular parts simplifies it and allows combination of the components that work well. In total, three state transition and three output prediction models were considered as presented in Sections \ref{sec:state_transition_problem} and \ref{sec:output_prediction_problem}, respectively. The policy learning problem was treated separately in Section \ref{sec:policy_problem} after finding a suitable plant approximation. 

From hereon, a common notation is adopted where the state and input vectors are defined as $\mathbf{x} = [\mathbf{q}, \, \mathbf{\dot{q}}]$ and $\mathbf{u} = \mathbf{\ddot{q}}$, respectively. Moreover, $f(\, \cdot \, ;\; \theta)$ represents a learnable function with inputs $(\cdot)$, such as a DNN, parameterised by $\theta$. The size of $\theta$ depends on the hyperparameters of the function. For example, for a DNN, the size of $\theta$ varies with the number of layers, the number of neurons per layer, the type of activation function, the type of layer, etc.

\subsection{State Transition Model Candidates}
\label{sec:state_transition_problem}
\subsubsection{NSSM}
\label{sec:dpc_mca_NSSM_state}

A Neural State Space Model (NSSM) is a type of black-box model inspired by the classic state space representation $x_{t+1} = Ax_t + Bu_t$ where the matrices $A$ and $B$ are substituted by learnable functions \cite{masti2018learning}. The specific form of the model that was considered is given by \eqref{eq:dpc_mca_NSSM_state}, where $f_1$ and $f_2$ are DNNs. 
\begin{equation}
  \label{eq:dpc_mca_NSSM_state}
  \mathbf{x}_{t+1} = f_1(\mathbf{x}_t; \; \theta_1) + f_2(\mathbf{u}; \; \theta_2)
\end{equation}

The learning objective of this model is to recursively predict the next state of the plant $(\mathbf{x}_{t+1})$ over horizons of length $n_p$ given an initial state $\mathbf{x}_{0}$ and a sequence of inputs $\mathcal{U}^j$. This learning objective can be achieved using the loss function \eqref{eq:dpc_mca_state_loss_nssm} which consists of one Mean Square Error (MSE) term, averaged over a total of $J$ training trajectories of length $n_p$. For each trajectory the MSE term is computed between the predictions of \eqref{eq:dpc_mca_NSSM_state} and the true trajectory contained in $\mathcal{X}^j$.
\begin{align}
  \label{eq:dpc_mca_state_loss_nssm}
  \min_{\theta_1, \, \theta_2} \quad & \frac{1}{J \, n_p}\sum_{j=1}^{J} \sum_{t=1}^{n_p} (\mathbf{x}_t - \mathcal{X}^j_t)^2            \\
  \textrm{s.t.} \quad                & \mathbf{x}_{0} = \mathcal{X}^j_0  \nonumber                                                   \\
                                     & \mathbf{x}_{t+1} = f_1(\mathbf{x}_t; \; \theta_1) + f_2(\mathcal{U}^j_t; \; \theta_2) \nonumber
\end{align}

\subsubsection{NODE 1 \& 2}
\label{sec:dpc_mca_NODE_state}

Neural Ordinary Differential Equation (NODE) models are well suited to tasks involving continuous transformations such as learning the dynamics of physical systems \cite{chen2018neural}. The analytical form of the state transition equation is an ODE, therefore, an NODE was a natural choice of approximation method. More specifically, two NODE formulations were considered as shown in \eqref{eq:dpc_mca_node_state_1} and \eqref{eq:dpc_mca_node_state_2}. The NODE 1 model in \eqref{eq:dpc_mca_node_state_1} uses the learnable function $f_1$ parameterised by $\theta_1$ to approximate the derivatives of the state variables. This learnable function can be integrated over time using an ODE solver and a suitable integration scheme. We used the adjoint method with Runge-Kutta integration. Alternatively, NODE 2 as defined in \eqref{eq:dpc_mca_node_state_2} was designed to be an exact implementation of the analytical state transition equation, where the states and inputs of the plant are derivatives of each other, e.g. $\int_{}^{}\ddot{\mathbf{q}} = \dot{\mathbf{q}}$ and $\int_{}^{}\dot{\mathbf{q}} = \mathbf{q}$. Thus, the function $I$ was set to a DNN with fixed parameters that acts as an identity map of suitable dimensions to enable integration of $\mathbf{\Ddot{q}}_{t}$ into $\mathbf{\dot{q}}_{t+1}$ and $\mathbf{q}_{t+1}$ accounting for the initial condition in $\mathbf{x}_0$ and using the forward Euler method. The integration time step was set to $0.01 \, \unit{\second}$ for both models.
\begin{align}
  \label{eq:dpc_mca_node_state_1}
  \dot{\mathbf{x}}(t) & = f_1(\mathbf{x}_t, \, \mathbf{u}_t, \, t;\; \theta_1), \, \mathbf{x}(0) = \mathbf{x}_0 \\
  & \Rightarrow \mathbf{x}_{T} = \int_{0}^{T}f_1(\mathbf{x}_t, \, \mathbf{u}_t, \, t;\; \theta_1) \; dt \nonumber \\
  \label{eq:dpc_mca_node_state_2}
  \dot{\mathbf{x}}(t) & = I(\mathbf{x}_t, \, \mathbf{u}_t), \, \mathbf{x}(0) = \mathbf{x}_0 \\
  & \Rightarrow \mathbf{x}_T = \int_{0}^{T}I(\mathbf{x}_t, \, \mathbf{u}_t) \; dt \nonumber
\end{align}

The training loss function for the first NODE model was identical in design and purpose to \eqref{eq:dpc_mca_state_loss_nssm}. The second NODE model does not require a loss function as it is an implementation of an exact equation with no trainable parameters.

\subsection{Output Model Candidates}
\label{sec:output_prediction_problem}

\subsubsection{NSSM}
\label{sec:dpc_mca_NSSM_output}

An NSSM model was also considered for the output prediction task. The specific form of the model is shown in \eqref{eq:dpc_mca_nssm_output} which follows the classic format $y_t = Cx_t + Du_t$ but instead of matrices, $C$ and $D$ are replaced by DNNs $f_3$ and $f_4$.
\begin{equation}
  \label{eq:dpc_mca_nssm_output}
  \mathbf{y}_{t+1} = f_3(\mathbf{x}_t; \; \theta_3) + f_4(\mathbf{u}; \; \theta_4)
\end{equation}
The learning objective of this model is to predict the next output of the plant $\mathbf{y}_{t+1}$ given its current state $\mathbf{x}_{t}$ and the next input to be applied $\mathcal{U}^j_t$. This learning objective can be achieved using the loss function \eqref{eq:dpc_mca_output_loss_nssm} which contains one MSE term computed between the predictions of \eqref{eq:dpc_mca_nssm_output} and the true trajectory contained in $\mathcal{Y}^j$.
\begin{align}
  \label{eq:dpc_mca_output_loss_nssm}
  \min_{\theta_3,\,\theta_4} \quad & \frac{1}{J \, n_p}\sum_{j=1}^{J} \sum_{t=1}^{n_p} (\mathbf{y}_t - \mathcal{Y}^j_t)^2            \\
  \textrm{s.t.} \quad              & \mathbf{x}_{0} = \mathcal{X}^j_0  \nonumber                                                   \\
                                   & \mathbf{x}_{t+1} = \text{pre-trained state transition model} \nonumber             \\
                                   & \mathbf{y}_{t+1} = f_3(\mathbf{x}_t; \; \theta_3) + f_4(\mathcal{U}_t^j; \; \theta_4) \nonumber
\end{align}
For this model there is no need to train on trajectories instead of individual points. Nevertheless, the trajectory approach was used for consistency across models and to enable combined training of the output model and a state transition model.

\subsubsection{Latent Space Model}
\label{sec:dpc_mca_latent_output}

The second output model candidate was a latent space model consisting of an encoder and a decoder as shown in \eqref{eq:dpc_mca_latent_output}. The encoder $f_3$, is a DNN that performs the mapping $f_3: \mathbf{y},\,\mathbf{x},\,\mathbf{u} \rightarrow \tilde{\mathbf{x}} \in \mathbb{R}^{l}$. The decoder $f_4$, performs the mapping $f_4: \mathbb{R}^{l} \rightarrow \mathbf{y}$. The latent space dimension $l$ is a model hyperparameter which was set to 12. 
\begin{align}
  \label{eq:dpc_mca_latent_output}
  \tilde{\mathbf{x}} & = f_3(\mathbf{y}_t,\,\mathbf{x},\,\mathbf{u}; \; \theta_3) \\
  \mathbf{y}_{t+1}   & = f_4(\tilde{\mathbf{x}}; \; \theta_4) \nonumber
\end{align}

For this model the learning objective is to recursively predict the next output $\mathbf{y}_{t+1}$ given the previous output $\mathbf{y}_t$, current state $\mathbf{x}_t$ and next input to be applied $\mathcal{U}^j_t$. Due to the recursive use of the output variable, training over trajectories was necessary and achieved with the loss function shown in \eqref{eq:dpc_mca_output_loss_latent}.
\begin{align}
  \label{eq:dpc_mca_output_loss_latent}
  \min_{\theta_3,\,\theta_4} \quad & \frac{1}{J \, n_p}\sum_{j=1}^{J} \sum_{t=1}^{n_p} (\mathbf{y}_t - \mathcal{Y}^j_t)^2                        \\
  \textrm{s.t.} \quad              & \mathbf{x}_{0} = \mathcal{X}^j_0  \nonumber                                                               \\
                                   & \mathbf{y}_{t=0} = \mathcal{Y}^j_0  \nonumber                                                               \\
                                   & \mathbf{x}_{t+1} = \text{pre-trained state transition model} \nonumber                         \\
                                   & \mathbf{y}_{t+1} = f_4(\;f_3(\mathbf{y}_t,\,\mathbf{x},\,\mathbf{u}; \; \theta_3)\;; \; \theta_4) \nonumber
\end{align}

\subsubsection{Mixed Analytical Model}
\label{sec:dpc_mca_analytical_output}

The third output prediction candidate was a mixed analytical model combining learnable functions and the exact equations of the plant. In practice, it is possible to implement the plant model as stated in \eqref{eq:analytical_plant} using a symbolic math toolbox like PyTorch. However, the exact implementation of the equations is computationally inefficient leading to prohibitive training and inference times. This problem was circumvented by substituting the slow operations by learnable functions as shown in Equations \eqref{eq:dpc_mca_nn_dFK} and \eqref{eq:dpc_mca_nn_ddFK} which are the replacements for \eqref{eq:dFK} and \eqref{eq:ddFK}, respectively. Note the one-to-one correspondence between the terms in \eqref{eq:dpc_mca_nn_dFK} and \eqref{eq:dFK} and between \eqref{eq:dpc_mca_nn_ddFK} and \eqref{eq:ddFK}, for example $f_3(g(\mathbf{\dot{q}}), \, \mathbb{F}; \; \theta_3) \leftrightarrow \mathbb{J}(\mathbb{F}(\mathbf{q}))\,\Dot{\mathbf{q}}$.
\begin{alignat}{2}
  \label{eq:dpc_mca_nn_dFK}
  \mathbb{\dot{F}}  & = f_3(g(\mathbf{\dot{q}}), \, \mathbb{F}; \; \theta_3)                                                        \\
  \label{eq:dpc_mca_nn_ddFK}
  \mathbb{\ddot{F}} & = f_4(g(\mathbf{\dot{q}}), \, \mathbb{\dot{F}}; \; \theta_4) + f_5(g(\mathbf{u}), \, \mathbb{F}; \; \theta_5)
\end{alignat}
In \eqref{eq:dpc_mca_nn_dFK} the function $g(\cdot)$ is a feature augmentation function set to $g(\lambda) = [\sin{(\lambda)}, \, \cos{(\lambda)}]$. Previous models attempt to learn \eqref{eq:analytical_plant} in an end-to-end fashion. In contrast, the learnable functions in this model target individual terms, thereby resulting in simpler regression problems.

The training loss function for this model is given by \eqref{eq:dpc_mca_output_loss_mixed} which contains two MSE terms. The first MSE term is the prediction error of $\mathbb{\dot{F}}$ as calculated from \eqref{eq:dpc_mca_nn_dFK} relative to $\mathcal{\dot{F}}^j$ which contains the real values of the derivative for the $j$--th training trajectory. Similarly, the second MSE term is the prediction error of $\mathbb{\ddot{F}}$ as calculated from \eqref{eq:dpc_mca_nn_ddFK} with respect to $\mathcal{\ddot{F}}^j$. Training over trajectories is also optional for this model because it does not include any recursive terms.
\begin{align}
  \label{eq:dpc_mca_output_loss_mixed}
  \min_{\theta_3, \, \theta_4, \, \theta_5} \quad & \frac{1}{J \, n_p}\sum_{j=1}^{J} \sum_{t=0}^{n_p-1} (\mathbb{\dot{F}}_t - \mathcal{\dot{F}}^j_t)^2 + (\mathbb{\ddot{F}}_t - \mathcal{\ddot{F}}^j_t)^2    \\
  \textrm{s.t.} \quad                             & \mathbf{x}_{0} = \mathcal{X}^j_0  \nonumber                                                                                                            \\
                                                  & \mathbf{x}_{t+1} = \text{pre-trained state transition model} \nonumber                                                                      \\
                                                  & \mathbb{F}_t = \prod^6_{i=0} T_i^{i-1}(q_{i_t}) \, \cdot \, T^6_{H}  \nonumber                                                                           \\
                                                  & \mathbb{\dot{F}}_t = f_3(g(\mathbf{\dot{q}}_t), \, \mathbb{F}_t; \; \theta_3)  \nonumber                                                                 \\
                                                  & \mathbb{\ddot{F}}_t = f_4(g(\mathbf{\dot{q}}_t), \, \mathbb{\dot{F}}_t; \; \theta_4) + f_5(g(\mathcal{U}_t^j), \, \mathbb{F}_t; \; \theta_5)   \nonumber
\end{align}

\subsection{Policy Model Design}
\label{sec:policy_problem}

After training a suitable plant model, the next step in the DPC design process is to build and train a policy model. Recall from Figure \ref{fig:method_explanation}, that the policy model is responsible for producing control actions that guide the plant towards a desired state or output given its current state. Herein, the policy model architecture was set to a bounded DNN as shown in \eqref{eq:dpc_mca_policy_network}. The learnable function $f_\pi$ maps the measured output $(\mathbf{y}_t)$, measured states $(\mathbf{x}_t)$ and target output $({\mathcal{Y}_t})$ to control actions $(\mathbf{u}_t)$ and an estimate of the plant states at the next time step $(\mathbf{\hat{x}}_{t+1})$. The saturation operator $sat_{lb}^{ub}(\cdot)$ is equivalent to $min(max(\cdot,\, lb),\, ub)$ where $lb$ and $ub$ are lower and upper bounds, respectively. For the version presented in \eqref{eq:dpc_mca_policy_network}, $lb = [\ddot{\mathbf{q}}^-,\, \mathbf{q}^-,\, \dot{\mathbf{q}}^-]$ and $ub = [\ddot{\mathbf{q}}^+,\, \mathbf{q}^+,\, \dot{\mathbf{q}}^+]$ which ensures that the policy respects all joint limits in Table \ref{tab:robotKinematics}.
\begin{equation}
  \label{eq:dpc_mca_policy_network}
  [\mathbf{u}_t, \, \mathbf{\hat{x}}_{t+1}] = sat_{lb}^{ub}\left(f_\pi(\mathbf{y}_t, \, \mathbf{x}_t, \, \mathcal{Y}_t; \; \theta_\pi)\right)
\end{equation}

The learning objective for the policy model is to recursively predict the control action $\mathbf{u}_t$ from the latest state and output measurements $\mathbf{x}_t$ and $\mathbf{y}_t$, such that the plant is regulated towards a reference output $\mathcal{Y}_t$. To achieve MPC-like performance, this recursion must result in stable and constraint-compliant state trajectories over long horizons. Thus, training happens using the closed loop setup shown in Figure \ref{fig:method_explanation} where the policy and pre-trained plant are connected in a cyclic manner. During training the policy attempts to regulate the pre-trained plant over pre-recorded trajectories of length $n_c$. The regulation performance of the policy is quantified using an MPC-inspired cost function such that the training problem is reduced to minimising the average value of that cost for a large set of training trajectories. Herein, the formulation shown in \eqref{eq:dpc_mca_policy_loss} was used. The value of $n_c$ used during training ultimately becomes the control horizon length that the policy learns to emulate. At inference time, the policy model will continue to behave as if the control horizon was $n_c$-steps long even if the value of $n_c$ is changed to 1. This is fundamentally different from MPC, where changing the length of the control horizon changes the behaviour of the controller and the solution of the underlying optimisation problem.
\begin{align}
  \label{eq:dpc_mca_policy_loss}
  \min_{\theta_4} \quad & \frac{1}{J \, n_c}\sum_{j=1}^{J} \sum_{t=0}^{n_c-1} \biggl((\mathbf{y}_{t+1} - \mathcal{Y}^j_t)^2 + (\mathbf{\hat{x}}_{t+1} - \mathbf{x}_{t+1})^2 \nonumber \\
  & \quad \quad \quad \quad \quad \quad \quad+ 0.1\,\mathbf{u}_t^2 + \mathbf{q}_t^2\biggl) \\
  \textrm{s.t.} \quad   & \mathbf{x}_{0} = \mathcal{X}^j_0                                                                                                                              \nonumber                       \\
                        & \mathbf{y}_{0} = \mathcal{Y}^j_0                                                                                                                              \nonumber                       \\
                        & \mathbf{u}^- \le \mathbf{u}_t \le \mathbf{u}^+                                                                                                                  \nonumber                       \\
                        & \mathbf{x}^- \le \mathbf{x}_t \le \mathbf{x}^+                                                                                                                  \nonumber                       \\
                        & \mathbf{x}_{t+1} = \text{pre-trained state transition model}                                                                                                    \nonumber                       \\
                        & \mathbf{y}_{t+1} = \text{pre-trained output prediction model}                                                                                                   \nonumber
\end{align}
The loss function \eqref{eq:dpc_mca_policy_loss} is made up of four terms averaged over $J$ $n_c$--long trajectories. The first term is the MSE between the achieved plant outputs $\mathbf{y}$ and the target references $\mathcal{Y}^j$. The second term is the MSE between the next state predictions $(\mathbf{\hat{x}}_{t+1})$ and the actual next states reached by the plant after applying the latest $\mathbf{u}_t$, as calculated by the state transition model. This term encourages the policy to learn the effect of the commanded input $\mathbf{u}_t$ on the plant state which makes the training more stable \cite{koyuncu2020novel}. The third term is the control action minimisation cost which discourages the policy from using large control inputs that could destabilize the system. Lastly, the fourth term is the washout cost -- a key part of every MCA. This cost encourages the policy to keep the end effector close to its starting position and to return to it upon large deviations such that future workspace availability is maximized.

\section{Training Data Generation}
\label{sec:datasets}
A DPC solution is typically trained from pre-recorded state and output trajectories. These recordings can be obtained from a real system with unknown dynamics, from simulation models, or from analytical equations. Given that the analytical form of the plant was known, we opted for the last option. In literature, the data generation problem from a known plant has been addressed using open- and closed-loop simulations. During open-loop simulations the plant is excited using input sequences such as sine waves, random walks or white noise \cite{drgovna2022differentiable}. Alternatively, during close-loop simulations the plant is regulated by a controller between starting and end states, or over pre-defined trajectories \cite{cavagnari1999neural, karg2019learning}. The advantage of the open-loop approach is that it does not require evaluating a control algorithm -- which can be computationally demanding. However, when the open-loop approach was tested on \eqref{eq:analytical_plant}, it was difficult to find input sequences that resulted in long trajectories without joint limit violations. Moreover, when such input sequences were found, they typically resulted in biased sampling favouring small regions around the starting positions. Thus, we adopted a hybrid approach where data from closed-loop simulations was used for training, and data from open-loop simulations was used for comparisons between models in scenarios that differed significantly from the examples seen during training and validation.

\subsection{Datasets From Closed-Loop Simulations}
\label{sec:dpc_mca_closed_loop_dataset}

The plant and policy learning problems call for a dataset $\mathcal{D}$ defined as in \eqref{eq:plant_dataset} which consists of $J$ trajectories of length $n_p$. Each of these trajectories corresponds to the system response when the control sequence $\mathcal{U}^j$ is applied starting at the initial conditions $\mathcal{X}_0^j$, $\mathcal{Y}_0^j$. In the dataset, $\mathcal{U}$ contains the input trajectories, $\mathcal{X}$ contains the state trajectories, $\mathcal{Y}$ contains the output trajectories, $\mathcal{F}$ contains the forward kinematics trajectories and $\mathcal{\dot{F}}$ and $\mathcal{\ddot{F}}$ contain the trajectories of the first and second forward kinematic derivatives.
\begin{align}
  \label{eq:plant_dataset}
  \mathcal{D} = {\{\,\{\mathcal{U}, \mathcal{X}, \mathcal{Y}, \mathcal{F}, \mathcal{\dot{F}}, \mathcal{\ddot{F}}}\}_{t=0, \, ..., \, n_p-1} \, \}^{j=1, \, ..., \, J}
\end{align}

For any given value of $n_p$, dataset $\mathcal{D}$ was generated by segmenting long closed-loop simulations of the plant into $n_p$-long chunks. The long closed-loop simulations were performed using the NMPC MCA proposed in \cite{arango2024model} and the downsampling approach proposed in \cite{gonzalez2024downsampling} to lower the computational burden of the simulations. The motion cueing scenario used during simulations was a fixed-wing aircraft flying task which was emulated in X-Plane 12 using the Cirrus SR-22 aircraft. In total, five 15-minute simulations were recorded where a human pilot was allowed to fly the aircraft using a joystick controller. No explicit instructions were followed during the flights, instead, efforts were made to include a wide range of operating conditions such as taxiing, take-off, landing, turbulence, stall manoeuvres, aggressive rolling, step inputs, and so on. During each flight, the approximate inertial signals experienced by the pilot were recorded using a simulated onboard IMU. These inertial signal recordings were then used to generate motion cueing trajectories for the serial robot simulator using the NMPC MCA at a control rate of $100\,\unit{\hertz}$. For $n_p = 256$, the resulting dataset $\mathcal{D}$ contained $1750$ example trajectories. For development, $\mathcal{D}$ was split into training $(\mathcal{D}_{train})$, validation $(\mathcal{D}_{val})$ and testing $(\mathcal{D}_{test})$ datasets using a 70:20:10 ratio and random shuffling before the split. As usual, the training dataset was used to minimise the model loss functions, the validation set was used to monitor over fitting and the testing set was used to evaluate the performance of models on unseen data.

\subsection{Datasets From Open-Loop Simulations}
\label{sec:dpc_mca_artificial_dataset}

The open-loop trajectory dataset used to test model generalisation beyond the training data was generated using the procedure described in Algorithm \ref{alg:dpc_mca_open_loop_dataset_generation}. The use of multiple signal types during simulations ensured a wide range of actuation frequencies. Meanwhile, initialising the plant to a random valid state at the start of each simulation increased coverage of the allowable states and outputs. During most simulations the plant was eventually driven beyond its state bounds. Therefore, only the bound-compliant portion of each simulation was used to add to the artificial dataset. Artificial trajectories generated from the open-loop method differed significantly to those generated from closed-loop simulations. Therefore, this dataset was a useful means to quantify how well a given model learnt the dynamics the analytical plant instead of the subset of dynamics displayed during closed-loop simulations for a particular task.

\begin{algorithm}[H]
  \caption{Open-loop dataset generation procedure}\label{alg:dpc_mca_open_loop_dataset_generation}
  \begin{algorithmic}[1]
    \STATE Initialise random number generators
    \FOR{$Signal$ $\in$ [step, walk, periodic, sine, splines]}
    \FOR{$i \gets 1$ to $6$}
    \FOR{$j \gets i$ to $6$}
    \FOR{$k \gets 1$ to $100$}
    \STATE Generate $\ddot{\mathbf{q}}$ profiles for joints i to j using $Signal$
    \STATE Initialise plant to random valid state
    \STATE Execute open-loop simulation $20 \unit{\second}$
    \STATE Identify first occurrence of state violation
    \STATE Trim output to time step before first violation
    \STATE Split simulated trajectory in $n_p$-long segments
    \STATE Append valid segments to $\mathcal{D}$
    \ENDFOR
    \ENDFOR
    \ENDFOR
    \ENDFOR
  \end{algorithmic}
\end{algorithm}

\section{Methodology}
\label{sec:methods}
Training of the new MCA was conducted in stages starting with the candidate state transition models, then the output prediciton models and lastly the policy model. First, the candidate models for the plant were trained using manually selected hyperparameters on the same dataset and with their respective loss functions. A curriculum training setup was used where each model was trained multiple times for an increasing horizon length. During the first training instance, the model's learnable parameters were randomly initialised. On subsequent intervals, training started from the best performing weights found during the previous interval. The sequence of horizon lengths (in steps) used during training was $[32,\, 64,\, 128,\, 256]$ for a maximum of $2.56 \, \unit{\second}$. The state transition models were trained and compared first. The output prediction models were then trained using the best state transition candidate.

Comparisons between plant models were made in terms of predictive accuracy, computational cost and ability to generalise to unexpected scenarios. Predictive accuracy was quantified in terms of RMSE with respect to reference signals in the validation and testing datasets. Computational cost was quantified as the average iteration time over 1000 inference calls on random samples from the training dataset. Lastly, ability to generalise to unexpected scenarios was quantified using the RMSE metric and the open-loop trajectory dataset. The most promising models identified from these tests were then used to train the policy model. 

After training the policy, the performance of the complete DPC-based MCA was assessed through experiments involving the analytical plant and a physics-based MuJoCo model. In experiments involving the analytical plant, we applied the new MCA to 10 pre-recorded motion cueing scenarios involving two different aircraft: the Cirrus SR-22 and the Cessna Citation X. The goal of the Cirrus SR-22 simulations (Case A), was to test if the new MCA could generalise to unseen scenarios of the aircraft that it was trained for. Alternatively, the goal of the Cessna Citation X simulations (Case B) was to test if the algorithm could also generalise to other aircraft with different handling dynamics. For each vehicle the performance of the new MCA was compared against that of the NMPC MCA presented in \cite{arango2024model} which uses the same joint-space plant formulation introduced in Section \ref{sec:analytical_plant}. Comparisons between the two algorithms were made in terms of the Performance Index $(PI)$ given by \eqref{eq:performance_metric} and computation time per iteration. For both test cases a Wilcoxon rank sum test was used to check if the distribution of performance indices across the 10 simulations differed when exact NMPC was used instead of the new MCA. The alpha level was set to 0.05.
\begin{align}
\label{eq:performance_metric}
PI = \sum_{k=1}^6 \left(\frac{{\|\mathbf{y}_v - \mathbf{y}_s\|}_2}{\sqrt{k}} - CC(\mathbf{y}_v,\, \mathbf{y}_s) \right)
\end{align}
$PI$ is a measure of similarity between the inertial signals experienced on the simulated vehicle ($\mathbf{y}_v$), and the inertial signals generated by the motion simulator ($\mathbf{y}_s$). The first and second terms of \eqref{eq:performance_metric} are the sum of RMSE and Pearson's Correlation Coefficient (CC), respectively. High similarity is characterised by a low value of RMSE and a high value of CC. Thus, a lower value of $PI$ indicates higher performance and the minimum value of $PI$ is $-6$, which corresponds to the case where RMSE is 0 and CC is 1 for all six components of the output signal $\mathbf{y}$.

For the second set of experiments involving the MuJoCo model in Figure \ref{fig:motionSimulator}, the new MCA was used to generate motion cues in real-time. During simulations a human user was allowed to pilot the Cirrus SR-22 using a joystick controller and inertial measurements gathered from X-Plane were sent via TCP to the MuJoCo simulation. In addition to the MCA, the MuJoCo environment included a computed torque controller to track the MCA commands. This hierarchy of controllers introduces unmodelled dynamics and uncertainty -- it is also the setup that would be expected when the MCA controls a real robot. Several simulations were conducted in this environment with the goal of assessing the stability, washout and constraint compliance properties of the new algorithm. 

\subsection{Training and Model Hyperparameters}

During experiments, training and model hyperparameters were manually selected after preliminary trials. The candidate plant models and policy model were trained using the RAdam optimiser. Each training session was allowed to run for up to 500 epochs per curriculum step for the plant and 3000 epochs for the policy. The training batch size was set to 256, the weight decay cost was set to 0.01, and the learning rate was set to 0.01 for the plant models and 0.001 for the policy model. Early stopping was enabled with the trigger criteria set to 150 epochs without validation loss improvements for the plant models, and 500 epochs for the policy. A simple ``reduce-on-plateau" learning rate schedule was used with a reduction factor of 0.5 triggered after 100 epochs of training loss stagnation. For the plant model candidates, all learnable functions were set to DNNs with two linear layers, 64 neurons per layer, GeLU activations and no bias. For the policy model $f_\pi$ was set to a bounded DNN with two linear layers, 128 neurons per layer, GeLU activations, no bias, and a ReLU clamp saturation method for the output. All random number generators were seeded with the same value at the start of each experiment.

\section{Results}
\label{sec:results}

\subsection{Plant Model Selection Experiments}
\label{sec:dpc_mca_preliminary_plant_selection_results}

Table \ref{tab:plant_selection_results} summarises the results for the state and output candidate selection experiments. From the state transition candidates that were considered, NODE 2 was the most competitive, it achieved the lowest RMSE values on both the validation and test datasets for all $n_p$ and it had the lowest average iteration time. Results from NODE 1 were close to those of NODE 2 and ultimately, both models outperformed the NSSM alternative by roughly three orders of magnitude in terms of predictive accuracy. The superiority of NODE 2 was expected because this model is an exact implementation of the analytical state transition equation and therefore its only source of error was integration. In contrast, NODE 1 was subject to regression error and its propagation through integration, and the NSSM model was at a disadvantage because it did not embed any form of implicit continuous transformation relationship like the NODE models. Therefore, NODE 2 was selected as the best state transition candidate and it was used during training of all output models. 

\begin{table*}[!ht]
\renewcommand{\arraystretch}{1.1}
\centering
\caption{Results from plant model selection experiments. Values in bold mark best performing model at each horizon length. }
\label{tab:plant_selection_results}
\begin{tabular}{c|c|ccc|ccc}
\multirow{2}{*}{Metric}        & \multirow{2}{*}{$n_p$} & \multicolumn{3}{c|}{State Transition Models}      & \multicolumn{3}{c}{Output Prediction Models}                         \\
                               &                        & NSSM  & NODE 1 & NODE 2         & NSSM           & Latent Space   & Mixed Analytical \\ \hline
\multirow{4}{*}{RMSE on validation set}  & 32                     & 1.905 & 0.002  & \textbf{0.001} & 0.320          & 0.327          & \textbf{0.186}   \\
                               & 64                     & 1.789 & 0.003  & \textbf{0.002} & 0.224          & 0.172          & \textbf{0.109}   \\
                               & 128                    & 2.100 & 0.010  & \textbf{0.003} & 0.250          & 0.157          & \textbf{0.143}   \\
                               & 256                    & 2.436 & 0.006  & \textbf{0.004} & 0.198          & \textbf{0.084} & 0.093            \\ \hline
\multirow{4}{*}{RMSE on test set} & 32                     & 1.919 & 0.003  & \textbf{0.001} & 0.349          & 0.368          & \textbf{0.232}   \\
                               & 64                     & 1.903 & 0.004  & \textbf{0.002} & 0.229          & 0.212          & \textbf{0.133}   \\
                               & 128                    & 1.770 & 0.005  & \textbf{0.003} & 0.221          & 0.100          & \textbf{0.095}   \\
                               & 256                    & 2.519 & 0.011  & \textbf{0.004} & 0.221          & \textbf{0.095} & 0.111            \\ \hline
\multirow{2}{*}{Average iteration time}             & 1   & 4.90E-04            & 7.73E-04
                & \textbf{2.54E-04}    & \textbf{5.37E-04} & 5.74E-04 & 0.002           \\
                               & 256                    & 0.063 & 0.136  & \textbf{0.023} & \textbf{0.098} & 0.099          & 0.368            \\ \hline
\multicolumn{1}{l|}{RMSE on open-loop dataset} & 256 & \multicolumn{1}{l}{} & \multicolumn{1}{l}{} & \multicolumn{1}{l|}{} & 20.612             & 19.094    & \textbf{10.158} \\ \hline
\end{tabular}
\end{table*}

Unlike the state transition models, the output prediction models were more evenly matched during experiments. As seen on Table \ref{tab:plant_selection_results}, the RMSE values achieved on both datasets were in the same order of magnitude for all $n_p$. In general, RMSE tended to decrease $n_p$ increased, indicating that the models benefited from training over longer horizons. At $n_p = 256$ the latent space model achieved the best performance on both datasets. The mixed analytical model followed closely, and it outperformed the latent space model on both datasets for $n_p < 256$. The NSSM model ranked in last place and its performance gap relative to the other models widened as $n_p$ increased, at $n_p = 256$ the model achieved RMSE values that were approximately double what the other two models achieved. In terms of computation speed, the NSSM and latent-space models were evenly matched with a slight advantage for the NSSM model. The mixed-analytical model was approximately three times slower due to its higher complexity. However, at $n_p=1$ all models were sufficiently fast to be evaluated at well over $100 \, \unit{\hertz}$ making them all suitable for real-time applications. The significant differentiator between models was the result of the comparison on the open-loop dataset which contained very different trajectories to those in the closed-loop dataset on which the models were trained. For this test, the analytical model outperformed the other models achieving an RMSE of $10.158$ compared to $20.612$ and $19.094$ for the NSSM and latent space models, respectively. The difference in performance was roughly two-fold. Previous results seemed to indicate that the latent space and mixed analytical models were closely matched in terms of predictive accuracy. However, this observation did not hold when the query data changed significantly from what the models were trained on, thereby showing that the mixed analytical model was the best approximation of the generalised plant dynamics. Due to this differentiating result, and despite being more computationally expensive than the NSSM and latent space alternatives, the mixed analytical model was selected as the most suitable candidate to be used  during policy training.

\subsection{Policy Experiments}

Having selected a state transition and output prediction model from the candidates, the policy network was trained until validation loss convergence using a frozen version of the pre-trained plant. During training, only the learnable parameters in $f_\pi$ were adjusted while the plant parameters remained constant. The trained policy was then evaluated on the control tasks described in Section \ref{sec:methods}.
\begin{figure}[!b]
    \centering
    \includegraphics[width=\linewidth]{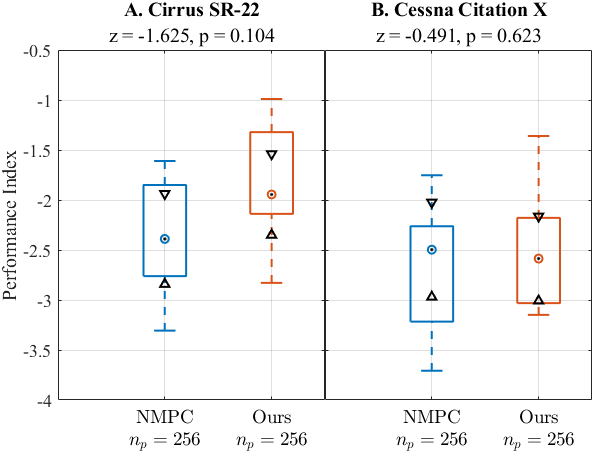}
    \caption{ Results from simulation experiments controlling analytical plant model. Box plots show the performance index distribution achieved by each MCA across the 10 scenarios considered for each aircraft. }
    \label{fig:analytical_plant_experiments}
\end{figure}
\begin{figure*}[!th]
    \centering
    \includegraphics[width=\textwidth]{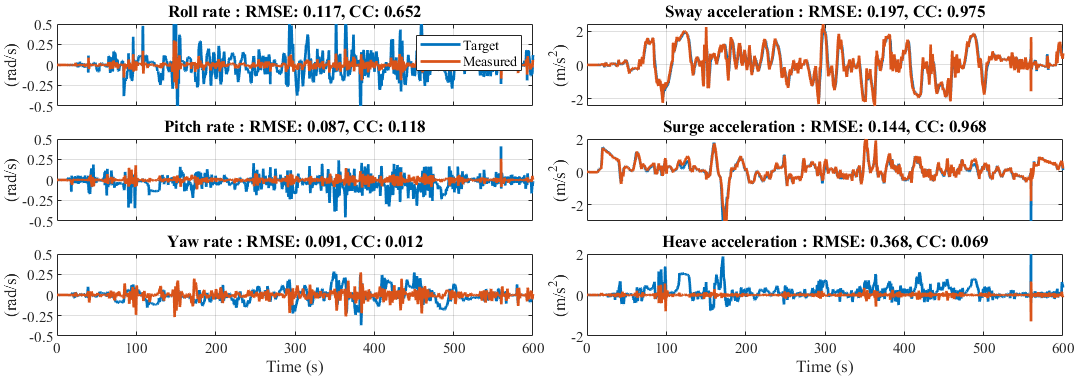}
    \caption{Target and achieved inertial signal profiles for 10 minute simulation in the real-time MuJoCo environment. The target profiles are the references for the MCA -- these references are scaled and gravity compensated versions of the IMU measurements gathered aboard the aircraft.}
    \label{fig:mujoco_experiments}
\end{figure*}
\begin{figure}[!b]
    \centering
    \includegraphics[width=\linewidth]{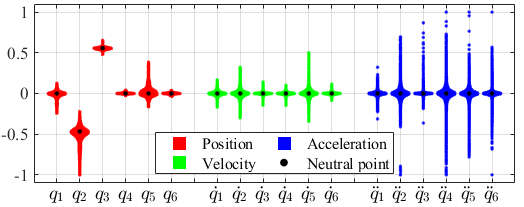}
    \caption{Swarm charts for joint positions, velocities and accelerations achieved during motion cueing simulation in Figure \ref{fig:mujoco_experiments}.}
    \label{fig:swarm_charts}
\end{figure}

Figure \ref{fig:analytical_plant_experiments} summarizes the results of the simulation experiments involving regulation of the analytical plant. For Case A, the distribution of $PI$ achieved by the new MCA heavily overlaps with the distribution achieved by exact NMPC, thus indicating similar performance. While exact NMPC achieved a lower median (-2.390) and lower minimum and maximum values than the new MCA (Mdn: -1.945), the Wilcoxon rank sum test indicated no significant difference in performance $(z = -1.625,\, p = 0.104)$ between the algorithms. Put simply, the new learning-based algorithm matches the performance of the NMPC controller it was designed to replace on unseen motion cueing scenarios. Moreover, results from Case B show that the same observation holds when both algorithms are used on a different aircraft. In test Case B, the median of the $PI$ distributions was lower for the new algorithm (-2.586) than for exact NMPC (-2.497) and the Wilcoxon rank sum test showed no difference in performance $(z = -0.491,\, p = 0.623)$. Therefore, the new MCA seems to generalise beyond the aircraft that it was trained for just as well as exact NMPC. In fact, both algorithms performed better during the Case B simulations than during Case A, likely because the Cessna Citation X is a larger aircraft with slower dynamics, and therefore, the reference profiles were not as challenging to match. 

Figure \ref{fig:mujoco_experiments} shows the output of a 10-minute simulation in the real-time MuJoCo environment where the Cirrus SR-22 aircraft was controlled with a joystick by a driver in the loop. The ``target" plots in the figure are the references that the MCA should track and the ``measured" plots are the values that were achieved during the simulation as calculated from the virtual IMU aboard the simulated motion platform. The RMSE and CC values reported in the title of each plot quantify the similarity between the target and measured plots. During this simulation and indeed, during most others, the new MCA prioritizes tracking of the sway and surge acceleration signals. In doing so, the algorithm makes ample use of tilt coordination which is why tracking along the roll and pitch DoF is compromised relative to sway and surge. Similarly, the algorithm struggles to achieve high CC values for yaw rate and heave acceleration tracking. The heave degree of freedom is particularly difficult to emulate for serial robot -based simulators due to the kinematic configuration of the joints which make it difficult to produce vertical movements without also incurring displacement in other directions. Nevertheless, the overall performance index for this simulation as calculated from \eqref{eq:performance_metric} is -1.79 which is within the confidence interval for the Cirrus SR-22 simulations using the new MCA in Figure \ref{fig:analytical_plant_experiments}. That is despite the simulation being 10 minutes long and using a physics-based model instead of the idealized plant. Moreover, during these and other simulations shown in the supplementary video of this publication, the new MCA exhibits uninterrupted stability, washout and constraint compliance with all joint limits. Figure \ref{fig:swarm_charts} shows swarm charts of the joint positions, velocities and accelerations achieved by the motion platform during the simulation in Figure \ref{fig:mujoco_experiments}. For convenience all points were normalised to the range $[-1, 1]$ using the limits in Table \ref{tab:robotKinematics}. Any point with a magnitude greater than unity is therefore in violation of the joint constraints but as seen on the plots, no violations occurred. For all joint variables the widest portion of the swarm chart is located around the neutral point which proves that the algorithm is effective at washout and control action minimisation.

Overall, these simulation experiments show that the performance of the new learning-based algorithm rival that of exact NMPC in terms of motion cueing quality. Across all simulations involved in Figure \ref{fig:analytical_plant_experiments}, the average computation time per iteration of exact NMPC was $0.809 \, \unit{\second}$. In contrast, the new learning-based algorithm was on average 400 times faster at $0.002 \, \unit{\second}$ per iteration with a maximum of $0.009 \, \unit{\second}$. Thus, the new algorithm is fast enough to be used in real-time at $100 \, \unit{\hertz}$ even when no measures have been taken to increase the execution speed of the code.

\section{Conclusion}
\label{sec:conclusion}

This study presented a novel learning-based MCA for serial robot-based motion platforms which leverages differentiable predictive control. The proposed MCA integrates the benefits of NMPC, including robust handling of constraints and kinematic complexities, with the computational efficiency of machine learning, allowing for rapid inference suitable for real-time applications. In Sections \ref{sec:dpc_mca} and \ref{sec:datasets} we provided extensive details of the model architectures and datasets that were considered during development, thereby laying the foundations for future research. Results from the plant model selection experiments in Section \ref{sec:dpc_mca_preliminary_plant_selection_results} showed that the best performing state and output prediction models were those that incorporated the analytical equations of the plant. Training the policy model with the selected candidates then proved successful. In simulation experiments, including generalization tests to new scenarios, a new vehicle and a new simulation environment, the learning-based algorithm maintained stability, washout, and constraint compliance while delivering motion cueing quality comparable to exact NMPC. However, unlike NMPC, our new algorithm can be used in real-time applications without resorting to reductions in horizon lengths, control rate or downsampling. This key advantage was achieved by shifting the computational burden to the offline training phase. During experiments, the average iteration time of our new algorithm was 400 times faster than the NMPC-based MCA used for comparisons. 

Our future work will focus on exploring the properties and use cases of the new approach. For instance, we are interested in gradient-based tuning methods to efficiently tailor the algorithm to new applications, on integration of learning-based methods to learn to control the system under disturbances and uncertainties, and on conducting experiments outside simulation.

\bibliographystyle{IEEEtran}
\bibliography{references}

\end{document}